\documentclass[letterpaper, 10 pt, journal, twoside]{IEEEtran}

\usepackage[utf8]{inputenc} % allow utf-8 input
\usepackage[T1]{fontenc}    % use 8-bit T1 fonts
\usepackage{graphicx}
\usepackage{amsmath}
\usepackage{amssymb}
\usepackage{booktabs}
\usepackage{stfloats}
\usepackage{epsfig}
\usepackage{svg}
\usepackage{subcaption,ragged2e}
\usepackage{caption} 
\usepackage[inline]{enumitem}
\usepackage{animate}
\usepackage{media9}
\usepackage{color}

\usepackage[pagebackref,breaklinks,colorlinks]{hyperref}

\usepackage[capitalize]{cleveref}
\crefname{section}{Sec.}{Secs.}
\Crefname{section}{Section}{Sections}
\Crefname{table}{Table}{Tables}
\crefname{table}{Tab.}{Tabs.}

\begin{document}

\title{Social-Pose: Enhancing Trajectory Prediction with Human Body Pose}

% \author{Yang Gao, Saeed Saadatnejad, Alexandre Alahi\\
% EPFL, Switzerland\\
% \texttt{firstname.lastname@epfl.ch} \\

% Saeed Saadatnejad\\
% EPFL, Switzerland\\
% % First line of institution2 address\\
% \texttt{saeed.saadatnejad@epfl.ch}\\
% % \And
% Alexandre Alahi\\
% EPFL, Switzerland\\
% \texttt{alexandre.alahi@epfl.ch}
% }

\author{Yang Gao, Saeed Saadatnejad, Alexandre Alahi,~\IEEEmembership{Member,~IEEE}
        % <-this % stops a space
\thanks{The authors are from the Visual Intelligence for Transportation (VITA) lab, École Polytechnique Fédérale de Lausanne (EPFL), 1015 Lausanne, Switzerland (e-mail:
yang.gao@epfl.ch).
This work was supported by Sportradar (Yang's Ph.D.), the European Union's Horizon 2020 research, innovation programme under the Marie Sklodowska-Curie grant agreement No 754354. }% <-this % stops a space
% \thanks{Manuscript received April 19, 2021; revised August 16, 2021.} % to be updated
}
\markboth{Journal of \LaTeX\ Class Files}% % to be updated
{Journal of \LaTeX\ Class Files}

\maketitle

%%%%%%%%% ABSTRACT
\begin{abstract}
Accurate human trajectory prediction is one of the most crucial tasks for autonomous driving, ensuring its safety. Yet, existing models often fail to fully leverage the visual cues that humans subconsciously communicate when navigating the space.
In this work, we study the benefits of predicting human trajectories using human body poses instead of solely their Cartesian space locations in time. 
We propose `Social-pose', an attention-based pose encoder that effectively captures the poses of all humans in a scene and their social relations.
Our method can be integrated into various trajectory prediction architectures. We have conducted extensive experiments on state-of-the-art models (based on LSTM, GAN, MLP, and Transformer), and showed improvements over all of them on synthetic (Joint Track Auto) and real (Human3.6M, Pedestrians and Cyclists in Road Traffic, and JRDB) datasets.
We also explored the advantages of using 2D versus 3D poses, as well as the effect of noisy poses and the application of our pose-based predictor in robot navigation scenarios. 
\end{abstract}


%%%%%%%%% BODY TEXT
\section{Introduction}
\label{sec:intro}

\IEEEPARstart{P}{redicting} future events is often considered an essential aspect of intelligence~\cite{bubic2010prediction}. This capability becomes critical in autonomous vehicles, where accurate predictions can help avoid accidents involving humans. For instance, consider a scenario where a pedestrian is about to cross the street. A non-predictive agent may only detect the pedestrian when they are directly in front, attempting to avoid a collision at the last moment. In contrast, a predictive agent can anticipate the pedestrian's actions several seconds ahead of time, making informed decisions on when to stop or proceed.

Trajectory prediction is usually defined as a sequence-to-sequence prediction task, with the goal of predicting future locations given past observations. It is commonly used in applications such as autonomous driving~\cite{zhao2022trajgat, saadatnejad2022socially, bahari2021injecting} and socially-aware robots~\cite{chen2019crowd, trautman2010unfreezing}. A key challenge in human trajectory prediction lies in its inherent stochasticity, as human behavior is influenced by free will. Nonetheless, people often provide subtle cues, such as their body language, gaze direction, and changes in speed or heading, that can signal their intentions.
For example, humans may turn their heads and shoulders before changing their walking directions; this visual cue cannot be captured with trajectories alone. 
Similarly, social interactions cannot be captured well if we ignore hand waves or head direction changes.
In this work, we propose to leverage the body signals that humans consciously or even subconsciously use to communicate their mobility patterns.

\begin{figure}[!t]
    \centering
    \includegraphics[scale=0.45]{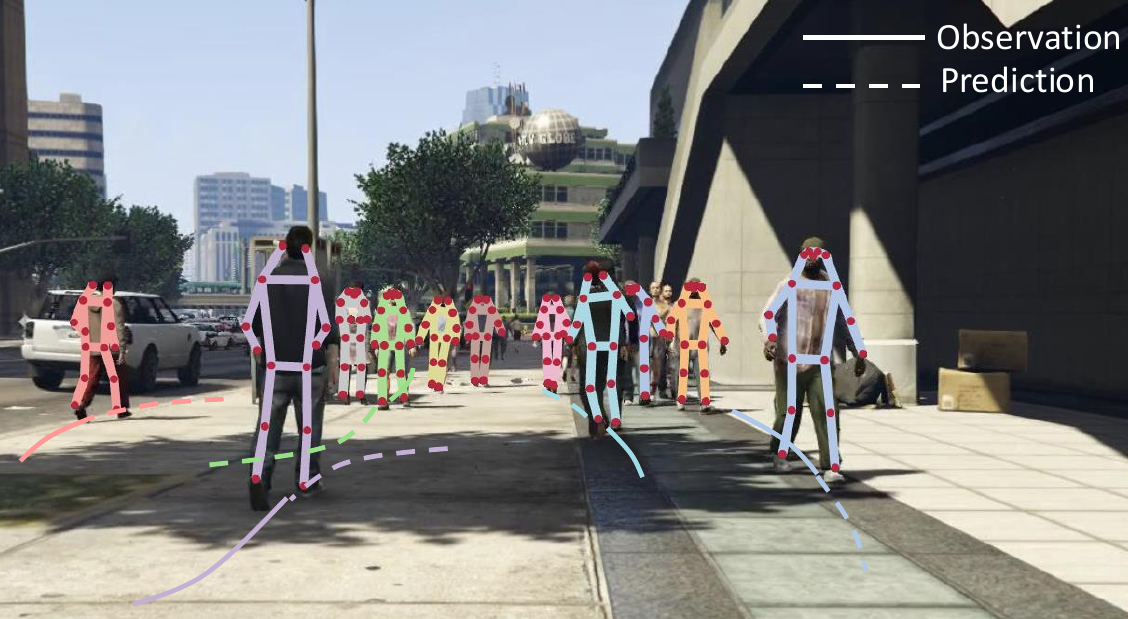}
    \caption{Given observed trajectories and pose keypoints of all agents, our model captures the spatiotemporal social interactions between them and predicts more accurate trajectories.}
    \label{fig:pull_figure}
\end{figure}

Pioneering works in human trajectory prediction mainly use humans' x-y locations in time as the input sequence~\cite{helbing1995social,alahi2016social}. However, humans are more than a point in space; they exhibit signals. As images contain numerous irrelevant details, we need to discover a better representation that captures the relevant cues. 
Some works~\cite{sadeghian2019sophie,dendorfer2021mg,huang2024fully} designed a decoupled module to learn scene representation and augment trajectory prediction. Inspired by that, this work aims to investigate the influence of a compact yet information-rich representation, namely body pose, on human trajectory prediction and provide a generic pose encoder to handle this augmented input efficiently.
The body pose consists of several keypoints of the person in 2D pixel coordinates or 3D world coordinates. By incorporating the sequence of observed poses along with the observed trajectories as input, the models predict future trajectories, as depicted in \Cref{fig:pull_figure}.

Several studies~\cite{saadatnejad2024socialtransmotion, gao2024multi, liang2019peeking, kress2022pose} have shown that using one specific pose encoder can help with one specific trajectory prediction model. However, a universal pose encoder
capable of integrating with diverse trajectory prediction networks remains an open challenge. Such an encoder, readily integrable with various existing architectures (e.g., LSTM-based, GAN-based, Transformer-based) with minimal modifications, is crucial as it would allow researchers in the motion prediction community to broadly leverage rich pose information without repeatedly re-engineering solutions. This facilitates quicker adoption across different models and applications.

To bridge this gap, we propose \textit{`Social-pose'}, a decoupled pose encoder that uses an attention-based encoder to capture rich information from human body poses. While its integration requires a minor architectural modification and retraining the model from scratch, its design is versatile, allowing it to be incorporated into various existing architectures (e.g., LSTM-based, GAN-based, Transformer-based).
By effectively leveraging pose information, our encoder can enhance prediction performance across different architectures, ensuring broader adaptability.
We conducted extensive experiments on state-of-the-art models with different architectures, including LSTM~\cite{alahi2016social}, GAN~\cite{gupta2018social}, MLP~\cite{xu2023eqmotion}, and Transformer~\cite{girgis2022latent} models, and observed improvements across all of them. We also questioned the necessity of using pose data from all individuals at all time steps and the need for 3D versus 2D poses, as only 2D poses may be available in some applications. Finally, we show that the pose encoder can be generalized to cyclists and is helpful in downstream robotic tasks to improve safety and efficiency. An in-depth analysis is provided in \Cref{sec:exp}.

To summarize, our contributions are three-fold:
\begin{enumerate}
    \item \textit{`Social-pose'}, a decoupled human body pose encoder, is introduced for trajectory prediction using an attention mechanism. This method can serve as a decoupled module for various trajectory predictors.
    \item An in-depth analysis is presented on the utilization of 3D/2D poses, including the impact of noisy or incomplete pose data in trajectory prediction scenarios, and its generalization to cyclists.
    \item The impact of pose-based predictors on downstream robot navigation tasks is demonstrated, highlighting their effect on safety and navigation speed.
\end{enumerate}

\section{Related Works}
\subsection{Human Trajectory Prediction}
Traditionally, human trajectory prediction is a sequence-to-sequence prediction task using a set of observed past positions as input and a set of predicted future positions as output. At an early stage, social force were used to tackle this task by modeling the attractive and repulsive forces among pedestrians~\cite{helbing1995social}. 
Later, Bayes Inference was utilized to predict human trajectories by modeling human-environment interaction~\cite{best2015bayesian}. 
Over time, data-driven methods have become increasingly prominent in the field~\cite{alahi2016social, chen2023vulnerable, uhlemann2024evaluating, liu2024egocentric, luan2025unified, rahimi2025sim}, with many studies constructing human-human interactions~\cite{alahi2016social, kothari2021human, monti2021dag, zhang2019sr} to improve predictions. For example, using hidden states from LSTM encoders to represent each agent’s motion dynamics and model interactions with neighboring pedestrians~\cite{alahi2016social}, or the directional grid for better social interaction modeling~\cite{kothari2021human}, and leveraging graph neural networks with nodes and edges to represent social dynamics~\cite{westny2023mtp, monti2021dag}.
Over the years, the research focus has expanded in trajectory prediction to encompass a broader range of social interactions~\cite{zhang2023bip,bhatt2023mpc,geng2023adaptive,wong2023msn,fu2024summary,zhou2024trajpred}, including human-context interactions~\cite{best2015bayesian, sun2022human} and human-vehicle interactions~\cite{li2024pedestrian,bhattacharyya2021euro, zhang2022learning}. Moreover, multimodality has been effectively modeled using various techniques, such as generative adversarial networks (GANs)~\cite{gupta2018social, hu2020collaborative, huang2019stgat}, Transformers~\cite{chen2023stochastic,yang2023long, zhong2023visual, girgis2022latent,yang2024multi}, diffusion models~\cite{gu2022trajdiffusion}, LLMs~\cite{lan2024traj}, and mixture density networks~\cite{lafage2024hierarchical}.

Transformers~\cite{vaswani2017attention} have been widely adopted for sequence modeling due to their ability to capture long-range dependencies and enable efficient parallel inference. As a result, this architecture has also gained significant traction in trajectory prediction tasks~\cite{yu2020spatio, giuliari2021transformer, li2022graph, yuan2021agentformer, girgis2022latent}. 
Most previous works have primarily relied on pedestrian x-y coordinates as input features. However, recent datasets providing 3D pose keypoints with more comprehensive information about pedestrian motion~\cite{fabbri2018learning, h36m_pami} have opened new possibilities. In this study, we exhaustively explore the potential benefits of incorporating these pose cues for different network architectures to enhance human trajectory prediction.

\subsection{Additional Inputs for Trajectory Prediction}
Multi-task learning is a credible way to share representations and make better use of complementary information for relevant tasks. Many pioneering works have shown that human trajectory prediction can also be improved by introducing extra associated tasks or information such as intention prediction~\cite{bouhsain2020pedestrian}, 2D/3D bounding-box prediction~\cite{bouhsain2020pedestrian, saadatnejad2022pedestrian}, 2D pose information~\cite{liang2019peeking}, and head pose forecasting~\cite{hasan2019forecasting}. However, in this study, we deviate from this category of work and instead focus on utilizing enriched input for trajectory prediction, exploring the potential benefits of incorporating 3D human pose information into the prediction process.

Human pose serves as a powerful indicator of human intentions, and recent advancements in pose estimation~\cite{kreiss2019pifpaf} have enabled the easy extraction of 2D poses from images.
While some works have explored using 2D pose keypoints for intention prediction~\cite{razali2021pedestrian,saleh2018intent} and trajectory prediction in the image/pixel space~\cite{yagi2018future,chen2020pedestrian}, our focus lies in trajectory prediction in camera/world coordinates, which holds more practical applications.  
One limitation of 2D keypoints lies in the potential loss of depth information, posing challenges in capturing spatial distances between agents accurately. In contrast, 3D keypoints do not suffer from this issue and have received significant attention in various applications, such as pose estimation~\cite{wandt2021canonpose}, pose prediction~\cite{saadatnejad2024unposed}, and pose tracking~\cite{reddy2021tessetrack}.

Recent studies~\cite{saadatnejad2024socialtransmotion, gao2024multi, kress2022pose} have explored the use of pose keypoints to enhance human trajectory prediction. However, these approaches often feature pose encoders that are tightly coupled with their specific network backbones, limiting their direct applicability as a general-purpose module for diverse architectures.

The goal of our work is to demonstrate that 3D/2D poses can be broadly beneficial across various architectures for predicting various pedestrian and cyclist trajectories, and also to explore the potential of poses in downstream navigation tasks.

\begin{figure*}[t]
    \centering
    
    \includegraphics[width= .94\textwidth]{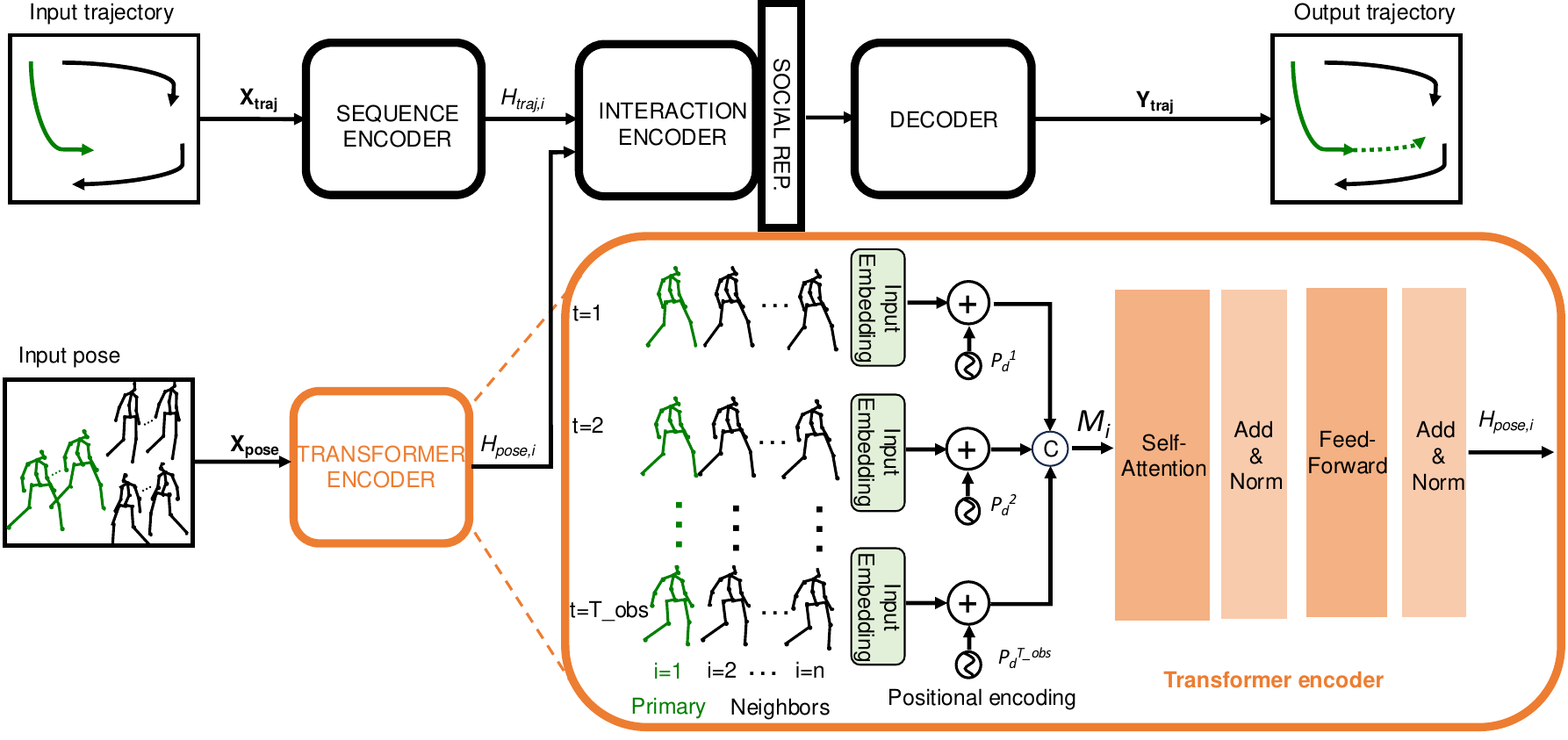}
    \caption{\textbf{Social-pose:} Our human pose encoder enhances trajectory prediction. It takes the sequence of observed poses of all people in the scene and generates a rich representation. This enriched information aids the trajectory decoder in predicting more accurate trajectories.}
    \label{fig:methods_overview}
\end{figure*}

\section{Method}
Our goal is to enhance existing trajectory prediction models by incorporating human poses as an additional input. To achieve this, we developed a decoupled pose encoder that learns a representation of pose cues and integrates it with the standard trajectory-encoded representation. The integration of this module requires a specific but minimal architectural modification and retraining the model end-to-end. As depicted in \Cref{fig:methods_overview}, our Transformer encoder serves as a decoupled module within the conventional trajectory prediction pipeline. 
The pose encoder, highlighted in orange, uses an attention-based Transformer encoder to capture spatiotemporal information from human body poses.

\subsection{Problem Formulation}
\label{chap:problem_formulation}
The task is to predict the future global trajectory coordinates. The observed time-steps are denoted by $t=1, ..., T_{obs}$ and the prediction timeframes are denoted by $t = T_{obs}+1, ..., T_{pred}$.  For pedestrian $i$ at time-step $t$, we denote the global trajectory coordinates as $\mathbf{x_{traj,i}^{t}} = (x_{i}^{t}, y_{i}^{t})$ and the local pose coordinates by $\mathbf{x_{pose,i}^{t}} = (x_{i,1}^{t},  y_{i,1}^{t}, z_{i,1}^{t}, \dots, x_{i,J}^{t},  y_{i,J}^{t}, z_{i,J}^{t})$, where $J$ is the number of body keypoints. The local pose represents the relative coordinates with respect to the pelvis joint. In a 3D pose, the \(x\) and \(y\) axes correspond to the horizontal dimensions, while the \(z\) axis represents the vertical dimension. In a 2D pose, we omit the \(z\) axis and use only the \(x\) and \(y\) axes to represent the coordinates in the image space.
We define $\mathbf{X_{traj,i}}$ and $\mathbf{X_{pose,i}}$ for the whole observations for pedestrian $i$. 
In a scene with $n$ pedestrians, the input of the network is denoted by $\mathbf{X} = \{\mathbf{X_{traj}}, \mathbf{X_{pose}}\}$, where $\mathbf{X_{traj}} = \{\mathbf{X_{traj,1}}, \dots, \mathbf{X_{traj,n}} \}$ and  $\mathbf{X_{pose}} = \{\mathbf{X_{pose,1}}, \dots, \mathbf{X_{pose,n}} \}$. The output of the network is denoted by $\mathbf{Y_{traj}} = \{\mathbf{Y_1}\}$, where $\mathbf{X}$ contains the observed trajectories and local pose, and $\mathbf{Y_1}$ contains the predicted future trajectory of the pedestrian that we are interested in (primary pedestrian).

\subsection{Pose Transformer Encoder}

To effectively extract pose features, an embedding layer converts the joint coordinates of all observed frames into input features for the Transformer encoder. Positional encoding is then applied to these embedded pose features to capture temporal information across different time steps. This encoding, implemented using sine and cosine functions similar to those used in natural language processing tasks~\cite{vaswani2017attention}, is mathematically defined for time-step $t$ as follows:

\begin{equation}
p_{d}^{t} = \left\{
  \begin{array}{rcc}
    \sin(\frac{t}{10000}^{d/D}), & \text{when d is even}\\
    \cos(\frac{t}{10000}^{d/D}), & \text{when d is odd}\\
  \end{array}
\right.,
\label{eq:positional_encoding}
\end{equation}
where $D$ is the feature dimension and $d$ is the dimension index. We follow the original formulation and use a maximum sequence length of 10000 to ensure the positional encodings span a wide range of frequencies. This choice maintains compatibility with standard Transformer implementations and does not affect computational efficiency, as only the actual number of time steps used in the data impacts the runtime. In practice, the model learns to focus within the effective temporal range while benefiting from the stable numerical properties of the full encoding spectrum. We denote the overall positional encoding at time-step $t$ as $p^t$ and we derive the intermediate embedding $M_{i}$ by adding positional features from $p^t$ to the embedded representation of $\mathbf{x^t_{pose,i}}$:

\begin{equation}
    M_{i} = (Emb(\mathbf{x^1_{pose,i}}) + p^1)  \oplus  \dots \oplus (Emb(\mathbf{x^{T_{obs}}_{pose,i}}) + p^{T_{obs}}).
\end{equation}

After incorporating positional encoding, the pose features undergo a series of transformations within the block. They pass through the self-attention module, followed by a residual connection. Subsequently, the features go through a feed-forward layer, and once again, a residual connection is applied. Then, the transformer encoder outputs the latent pose representations:
\begin{equation}
    H_{pose,i} = \mathbf{Enc}(M_{i}),
\end{equation}
where $H_{pose,i}$ is the learned representation of the pose of the $i$-th agent. It is then concatenated with the representation of the trajectory of the same agent:
\begin{equation}
    H_i = H_{pose,i} \oplus H_{traj,i}.
    \label{eq:concatenation}
\end{equation}

\begin{figure}[t]
    \centering
    \includegraphics[width=\linewidth]
    {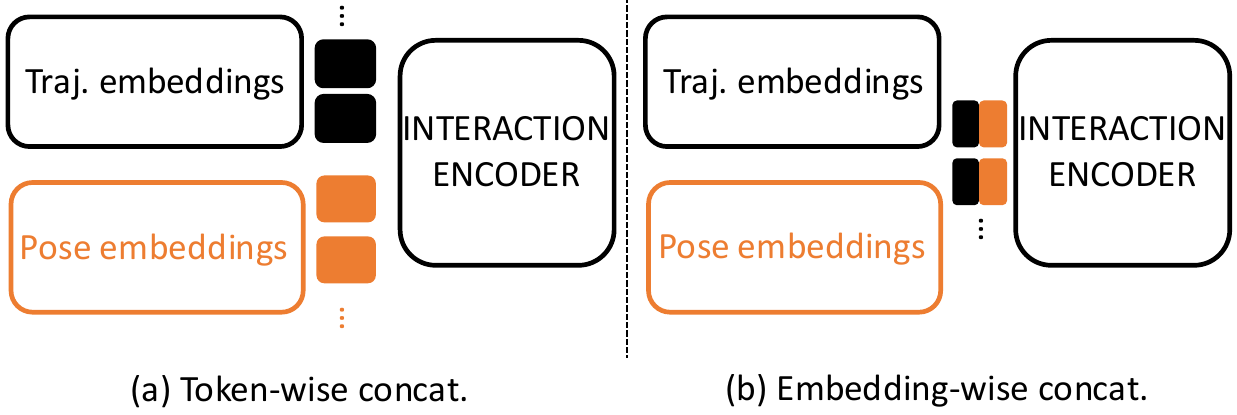}
    \caption{Comparison between work~\cite{saadatnejad2024socialtransmotion} and our Social-Pose. \cite{saadatnejad2024socialtransmotion} (a) fuses embeddings in token-wise concatenation. Our Social-Pose (b) uses embedding-wise concatenation for better compatibility with different trajectory predictors.}
    \label{fig:fusion_comparison}
\end{figure}

This process is executed for all pedestrians independently. The learned representations are then fed into the interaction encoder to extract interactions between agents.
Finally, the decoder is responsible for predicting the trajectory of the primary pedestrian.
\Cref{fig:methods_overview} visually illustrates the entire pipeline and the decoupled pose encoder that processes both the trajectory and local pose keypoints of pedestrians. \Cref{fig:fusion_comparison} illustrates the key architectural difference between our approach and existing work that also uses a transformer for pose encoding~\cite{saadatnejad2024socialtransmotion}. By employing embedding-wise concatenation, our pose encoder functions as a versatile module that can be readily integrated with different trajectory prediction backbones.

The specific architectures for the sequence encoder, interaction encoder, and decoder are adopted from the respective baseline models, and their implementation details, along with our pose encoder, are further elaborated in \Cref{sec:implementation}.

\section{Experiments}\label{sec:exp}
This section starts by introducing the datasets, evaluation metrics, baselines, and implementation details. We then present extensive quantitative and qualitative results, followed by analysis on the pose integration and extension in pixel-space trajectory prediction, as well as the generalization to cyclist trajectory prediction. Finally, we explore the application of our pose encoder in robot navigation scenarios.

\subsection{Datasets}
Trajnet++~\cite{kothari2021human} is a dataset for training and evaluating human trajectory prediction models in crowds. It offers a balanced dataset with diverse types of trajectories, making it valuable for trajectory prediction research.
However, since the original Trajnet++ dataset lacks pose information, we conducted our experiments on four publicly available datasets: JTA~\cite{fabbri2018learning}, Human3.6M~\cite{h36m_pami}, Pedestrians and Cyclists in Road Traffic~\cite{kress2022pose}, and JRDB~\cite{vendrow2023jrdb}, which provide 3D pose keypoints or 2D pose keypoints. Leveraging the Trajnet++ toolbox, we effectively categorized and balanced these datasets based on four trajectory types: static trajectories, linear trajectories, interaction trajectories, and other trajectories.
Following the Trajnet++ benchmark convention, we predict 12 future time steps given 9 past time steps at a frame rate of 2.5 fps on the JTA, Human3.6M, and JRDB datasets. Due to sequence length limitations in the Pedestrians and Cyclists in Road Traffic dataset, we instead predict 12 future frames from 4 historical frames at 5 fps. Since our focus is on leveraging informative 3D pose information, we exclude the JRDB dataset from our main experiments, as it does not provide ground-truth 3D pose annotations.

\subsubsection{JTA~\cite{fabbri2018learning}}
The JTA dataset is a large-scale synthetic dataset containing 256 video clips for training, 128 for validation, and 128 for testing, with approximately 10 million 3D/2D keypoint annotations in total. To capture accurate global trajectory information, only static-camera video clips are used. Our models are trained on 206 video clips and validated and tested on 10 and 12 static-camera clips, respectively. After pre-processing using Trajnet++, the training split comprises over 88k scenes, while the test split includes over 5k scenes, ensuring reliable results in interactive scenarios.

\subsubsection{Human3.6M~\cite{h36m_pami}} 
The Human3.6M dataset is a real-world dataset containing 3.6 million 3D pose annotations, featuring single-agent scenarios without pedestrian interactions. In our experiments, we focus on global body movement by including only walk-related activities (walking, walktogether, walkdog) and excluding other activities. 
Subsequently, we use S1, S5, S6, S7, and S8 for training, S11 for validation, and S9 for test.
In all our experiments, we considered 17 joints, the same as~\cite{martinez2017simple}.

\subsubsection{Pedestrians and Cyclists in Road Traffic~\cite{kress2022pose} (Urban dataset)}
It is a real-world dataset containing more than $2000$ trajectories of pedestrians and cyclists with 3D body poses recorded in urban traffic environments. 
It is specifically designed for single-person scenarios in urban traffic and has gained attention for research in autonomous driving.
The test set provides more than 50k scenes, enabling comprehensive evaluations of the models.

\subsubsection{JRDB~\cite{vendrow2023jrdb}} This real-world dataset offers a diverse collection of pedestrian trajectories and 2D body poses. Since the official test set is hidden, we use only the fixed-camera scenarios from the official training split for this experiment. As a result, we have 819 test scenes and 7649 training scenes, each with ground truth trajectories and 2D poses.

\subsection{Metrics}
We evaluate the models in terms of Average Displacement Error (ADE), Final Displacement Error (FDE), and Average Specific Weighted Average Euclidean Error (ASWAEE)~\cite{kress2022pose}:
\begin{enumerate}
    \item ADE: the average L2 displacement error between the predicted location and the real location of the pedestrian across all prediction timeframes; 
    \item FDE: the L2 displacement error between the predicted location and the real location of the final prediction timeframe;
    \item ASWAEE: the average displacement error per second for some specific time points; Following~\cite{kress2022pose}, we compute it for five timeframes: [t=0.44s, t=0.96s, t=1.48s, t=2.00s, t=2.52s].
\end{enumerate}

\subsection{Baselines}
To ensure a comprehensive evaluation, we selected a diverse set of baselines, including both interaction-aware and interaction-agnostic models with different architectures such as LSTM, GAN, MLP, and Transformer. Following the Trajnet++ leaderboard~\cite{kothari2021human}, we carefully chose baseline models that perform well in terms of accuracy. We integrate our pose encoder into the following baselines and evaluate their performance. 

\begin{itemize}
    \item Autobots~\cite{girgis2022latent}: a Transformer model that leverages temporal attention and spatial attention modules to model social interactions.
    
    \item EqMotion~\cite{xu2023eqmotion}: an MLP-based model that learns Euclidean geometric transformation to model the motion equivariance and interaction invariance.
    
    \item Social-LSTM~\cite{alahi2016social}: an LSTM model that utilizes social pooling layers based on hidden states to model interactions between agents.
    
    \item Vanilla-LSTM: a basic LSTM model with an interaction-agnostic encoder.
    \item Social-GAN~\cite{gupta2018social}: a Generative Adversarial Network that utilizes a max-pooling function to model social interactions.
\end{itemize}

Additionally, we report the performances of four extra baselines: Trajectory Transformer~\cite{giuliari2021transformer}, Directional-LSTM~\cite{kothari2021human}, Dir-Social-LSTM~\cite{kothari2021human}, Directional-GAN~\cite{kothari2021human}, and Social-Transmotion~\cite{saadatnejad2024socialtransmotion}.

\subsection{Implementation Details}
\label{sec:implementation}
We utilized the default architectures for all baselines in sequence encoding, interaction modeling, and trajectory decoding. For LSTM-based models like Social-LSTM~\cite{alahi2016social} and for the GAN-based baselines, an LSTM architecture serves as both the sequence encoder and decoder. In these models, interactions are captured using a pooling mechanism; Social-LSTM uses a social-pooling layer, while the GAN models employ a general pooling module combined with an adversarial discriminator to enforce socially compliant behaviors. In contrast, EqMotion~\cite{xu2023eqmotion} employs MLP-based encoders and decoders, leveraging a Graph Neural Network (GNN) to explicitly model interactions. Finally, Autobots~\cite{girgis2022latent} utilizes a temporal attention mechanism for sequence encoding, spatial attention for interaction modeling, and a standard transformer decoder to generate the output trajectories.

For all four LSTMs (Vanilla, Social~\cite{alahi2016social}, Directional~\cite{kothari2021human}, Dir-social~\cite{kothari2021human}) and two GANs (Social~\cite{gupta2018social}, Directional~\cite{kothari2021human}) networks, we set the embedding dimension to 64 to encode the displacement of global positions, and the pooling dimension of the interaction encoder to 256. After incorporating pose information, we double the interaction encoder’s dimension to enable the model to capture both trajectory and pose interactions.
The hidden dimension of both the LSTM encoder and decoder is consistently set to 128. 
For optimization, the Adam optimizer~\cite{kingma2014adam} was used, setting the initial learning rate to 0.001 and employing a scheduler to decay the learning rate every 10 epochs.

For the Transformer-based architecture~\cite{girgis2022latent}, we use the same settings for both the baseline model and the model augmented with pose information. Specifically, we use two layers for both the encoder and decoder. Each multi-head attention module consists of 16 heads, and the batch size is set to 64. The hidden dimension is fixed at 128 throughout the entire model. During training, we set the initial learning rate to $7.5 \times 10^{-4}$ and apply a decay factor of 0.5 every 10 epochs.
Our Transformer encoder processes $n$ pedestrians along the batch dimension, enabling it to handle scenes with varying numbers of pedestrians. 
We followed the original loss implementations for all methods: the joint loss for Autobots~\cite{girgis2022latent}, which combines negative log-likelihood, Kullback–Leibler divergence, and mean squared error (MSE); the auxiliary loss for GANs~\cite{gupta2018social}, which includes adversarial and MSE terms; and the standard MSE loss for the remaining methods. By maintaining consistent settings for both the baseline and the pose-augmented model, we ensure a fair comparison between the two approaches and enable a more meaningful evaluation of the impact of pose information on trajectory prediction.

\subsection{Quantitative Results}

\begin{table*}[!t]

    \centering
    \bgroup
    \def\arraystretch{1.5}%
    
    \begin{tabular}{llccc}
    \toprule

        \textbf{Model} & \textbf{Input Modality} & \textbf{JTA}~\cite{fabbri2018learning} & \textbf{Human3.6M}~\cite{h36m_pami} & \textbf{Urban}~\cite{kress2022pose}\\
        & & \textbf{ADE/FDE} & \textbf{ADE/FDE} & \textbf{ADE/FDE}  \\
        \midrule

        Directional-GAN~\cite{kothari2021human}& Traj & 1.83/4.33 & 0.62/1.02 & 0.60/1.09\\
        Trajectory Transformer~\cite{giuliari2021transformer}& Traj & 1.56/3.54 & 0.85/1.36  & 0.60/1.11\\
        Directional-LSTM~\cite{kothari2021human}& Traj & 1.37/3.06 & 0.60/0.99 & 0.58/1.06 \\
        Dir-social-LSTM~\cite{kothari2021human}& Traj & 1.23/2.59 & 0.58/0.95 & 0.58/1.06 \\
        Social-Transmotion~\cite{saadatnejad2024socialtransmotion}& Traj + 3D Pose & 0.94/1.94 & 0.54/0.89 & \textbf{0.57}/1.04 \\
        \midrule
        Vanilla-LSTM & Traj & 1.44/3.25 & 0.58/0.95 & 0.58/1.06 \\
        Vanilla-LSTM + Pose encoder (ours)& Traj + 3D Pose & 1.31/3.00 & 0.52/0.84 & \textbf{0.57}/1.04\\
        \midrule
        Social-GAN~\cite{gupta2018social}& Traj & 1.66/3.76 & 0.56/0.90 & 0.60/1.08\\
        Social-GAN + Pose encoder (ours)& Traj + 3D Pose  & 1.49/3.37 & 0.53/0.88 & 0.59/1.08 \\
        \midrule
        Social-LSTM~\cite{alahi2016social}& Traj & 1.21/2.54 & 0.60/0.93 & 0.58/1.06 \\
        Social-LSTM + Pose encoder (ours)& Traj + 3D Pose & 1.11/2.34 & 0.53/0.86 & \textbf{0.57}/1.04 \\
        \midrule
        EqMotion~\cite{xu2023eqmotion}& Traj & 1.13/2.39 & 0.51/0.81 & 0.58/1.05 \\
        EqMotion + Pose encoder (ours)& Traj + 3D Pose & 1.07/2.28 & \textbf{0.48}/0.79 & \textbf{0.57}/1.05 \\
        \midrule
        Autobots~\cite{girgis2022latent}& Traj & 1.20/2.70 & 0.55/0.84 & 0.58/1.04 \\
        Autobots + Pose encoder (ours)& Traj + 3D Pose & \textbf{0.90}/\textbf{1.91} & 0.53/\textbf{0.74} & \textbf{0.57}/\textbf{1.03} \\

        \bottomrule
    \end{tabular}
    \egroup
    \caption{Quantitative results on the three datasets with Ground Truth 3D pose. ADE and FDE are reported in meters.}
    \label{tab:main_table}
\end{table*}

\Cref{tab:main_table} provides comprehensive quantitative results on the JTA~\cite{fabbri2018learning} dataset, the Human3.6M~\cite{h36m_pami} dataset, and the Urban dataset~\cite{kress2022pose}, comparing the performance of baseline models with and without our proposed pose encoder. The results with consistent improvement on different architectures demonstrate the success and universality of our framework.

\begin{figure*}
 
  \begin{subfigure}{\linewidth}
  \centering
    \includegraphics[width=.90\textwidth]{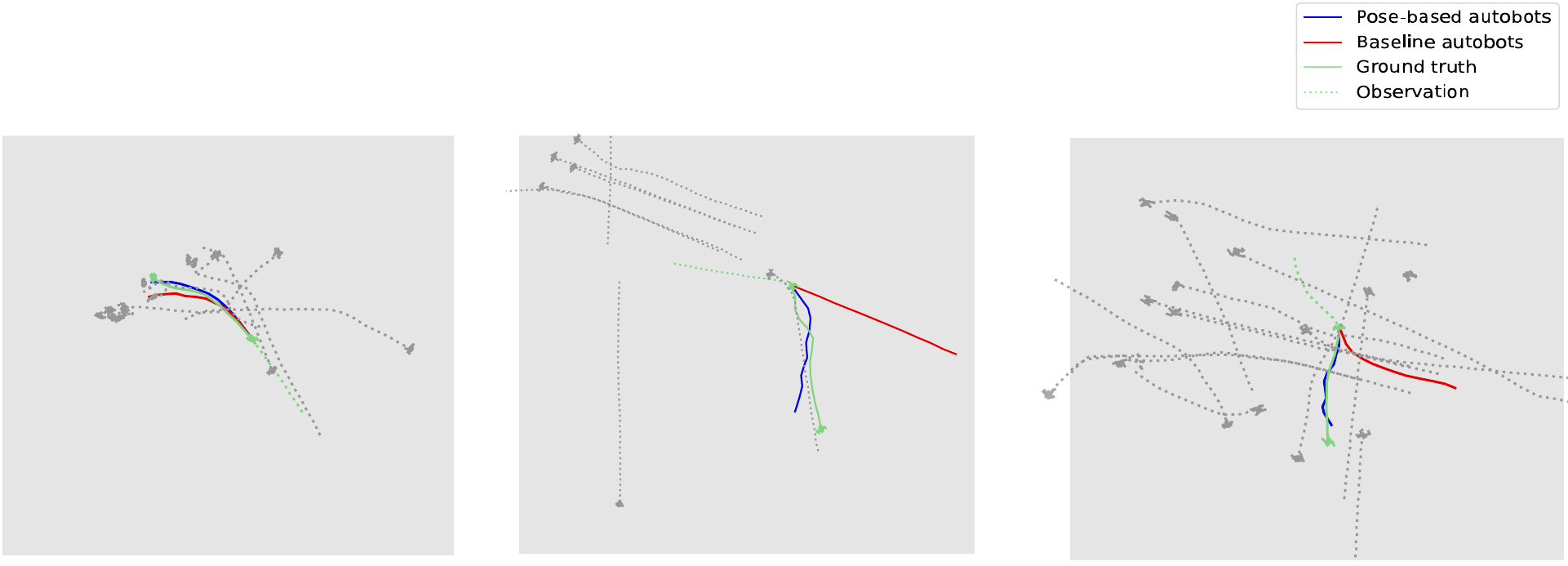}
    \label{fig:aa}
  \end{subfigure}
  \hfill
  \caption{Qualitative examples on the JTA~\cite{fabbri2018learning} dataset. Each example depicts pedestrian trajectories within a specific scene. For the primary agent, the ground truth is shown in green, the baseline model’s prediction in red, and the pose-based model’s prediction in blue. All other agents are represented in gray. The pose of the last observed frame is also visualized, as it indicates walking direction and body rotation.}

  \label{fig:vis_jta}
\end{figure*}

On the JTA~\cite{fabbri2018learning} dataset, incorporating pose information consistently improves ADE and FDE metrics across all evaluated baseline architectures (LSTM, GAN, MLP, and Transformer), with gains of up to 25\% and 29\%, respectively. This improvement can be attributed to the model’s ability to predict more accurate turning angles after introducing pose, as demonstrated in the qualitative results presented in \Cref{fig:vis_jta}. 

Furthermore, the strength of our decoupled pose encoder design is highlighted when compared to existing state-of-the-art methods that also utilize pose information.
Notably, when our Social-Pose is integrated with the Autobots baseline, the performance surpasses Social-Transmotion~\cite{saadatnejad2024socialtransmotion}, a state-of-the-art model also trained with trajectories and 3D poses. This demonstrates that our Social-Pose framework not only enhances various architectures but also empowers them to achieve or exceed state-of-the-art performance, validating its efficacy in improving trajectory prediction.

Similarly, adding pose information enhances performance on the Human3.6M~\cite{h36m_pami} dataset, with all pose-based models achieving lower ADE/FDE than their baseline counterparts, which validates the generalizability of our pose encoder on different architectures. Notably, the baseline EqMotion~\cite{xu2023eqmotion} outperforms the baseline Autobots~\cite{girgis2022latent} in ADE, likely because MLPs are effective on smaller datasets. Since the Human3.6M~\cite{h36m_pami} dataset only involves single-pedestrian scenarios without interactions, all improvements are due to the pose information of the primary pedestrian.

On the Urban~\cite{kress2022pose} dataset, the performance differences across methods are relatively small due to the prediction horizon being 50\% shorter compared to the JTA/Human3.6M datasets, and the absence of neighboring pedestrians for interaction modeling. Nonetheless, all four types of architectures show consistent improvement after integrating our pose encoder.

We select Autobots as the primary model for subsequent experiments due to its superior performance.

\begin{figure}[!t]
  \centering
  \includegraphics[width=.3\textwidth]{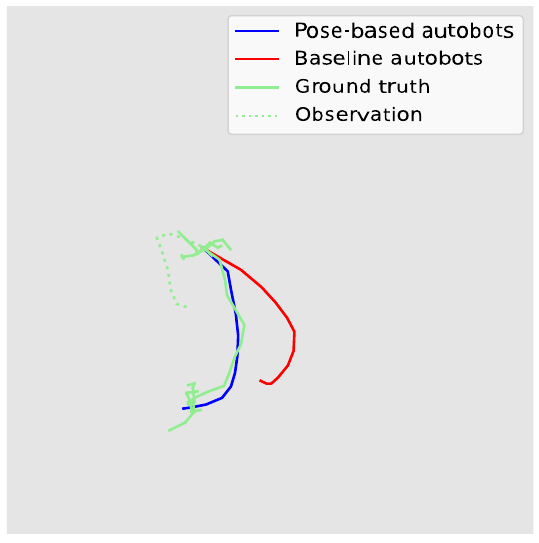}

  \caption{Qualitative examples on the Human3.6M~\cite{h36m_pami} dataset. For the primary agent, the ground truth is shown in green, the model's prediction in red, and the pose-based model's prediction in blue. All other agents are depicted in gray.}
  \label{fig:vis_h36m}
\end{figure}

\subsection{Qualitative Results}
\Cref{fig:vis_jta} shows visual comparisons between the original Autobots model and its pose-based version. 
The visualizations demonstrate that incorporating body rotation improves the prediction of directional changes, allowing for more complex trajectories beyond simple linear paths. Furthermore, the model shows better handling of social interactions, as illustrated in the right-most figure. Without pose cues, the model incorrectly predicts a left turn. However, with pose information, it accurately captures the body rotations of all agents, resulting in more precise trajectory predictions.

\Cref{fig:vis_h36m} presents a qualitative example from the Human3.6M~\cite{h36m_pami} dataset, comparing the performance of original Autobots and its pose-based version. The visualization shows that pose-based models generate trajectories that more closely align with the ground truth than its baseline counterparts.

\subsection{Pose Integration Analysis}
In this section, we present more analysis of incorporating human body pose into trajectory prediction. All experiments are conducted on the JTA~\cite{fabbri2018learning} dataset as it offers a large and diverse set of samples for thorough evaluation.

\subsubsection{Computational Cost}
To assess the practicality of the pose encoder, it is essential to evaluate its computational cost overhead. We report inference speed on the full test set, which contains over 5,000 samples, by computing the average and standard deviation over five runs on a single NVIDIA RTX 3090 GPU with a batch size of 1. As shown in \Cref{tab:computatoinal_cost}, the inference time for Autobots + 3D Pose is only slightly higher (around 2\%) compared to Autobots without pose. This minor overhead is negligible, especially given the significant improvements in ADE/FDE (approximately 25\%) achieved by incorporating pose information with our encoder.

\begin{table}[!t]
    \centering
    \bgroup
    \def\arraystretch{1.5}%
    \begin{tabular}{lcc}
    \toprule
        \textbf{Models} &  \textbf{Inference time} & \textbf{ADE/FDE}\\
        \midrule
        Autobots~\cite{girgis2022latent} & $\mathbf{7.56\pm 0.05 }$ miliseconds  & 1.20/2.70\\
        Autobots + 3D Pose & $7.70\pm 0.07 $ miliseconds & \textbf{0.90/1.91}\\

         \bottomrule
         
    \end{tabular}
    \egroup
    \caption{Computational cost comparison when adding the pose encoder on the JTA~\cite{fabbri2018learning} dataset.}
    \label{tab:computatoinal_cost}
\end{table}

\subsubsection{Attention Maps}

\begin{figure*}[!t]
    \centering
    \includegraphics[width=.8\textwidth]{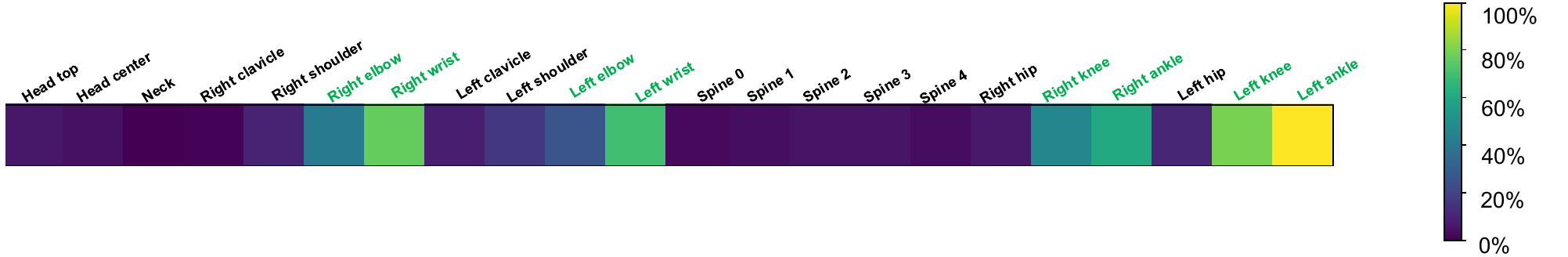}
    \caption{Attention map for joints. Spatially, the arms and legs are more significant in trajectory prediction.}
    \label{fig:general_att_map}
\end{figure*}

The attention mechanism in our pose encoder offers valuable insights into the temporal and spatial factors influencing the model's decision-making process. Specifically, we visualize the spatial attention map by averaging the attention weights across all frames for each joint. To avoid bias toward specific samples, we calculate attention scores across the entire test set of over 5,000 samples. This allows us to identify which frames and pose keypoints contribute most significantly to improving trajectory prediction.
 \Cref{fig:general_att_map} highlights some specific keypoints the model focuses on most, such as the ankles, knees, wrists, and elbows, underscoring the importance of arms and legs in guiding the model’s predictions. As shown in \Cref{tab:selected_joints}, utilizing only the 8 highest-scoring joints yields a notable 23.3\%/28.1\% improvement over the baseline. However, this performance is slightly inferior to that achieved with full-body joints, indicating that even joints with lower attention scores contribute valuable information.

\begin{table}[!t]  
    \centering
    \bgroup
    \def\arraystretch{1.5}%
    \begin{tabular}{lcc}
    \toprule
        \textbf{Models} &  \textbf{ADE/FDE}  & \textbf{(gain)} \\

        \hline

         Autobots + 3D Pose (full-body joints) & \textbf{0.90/1.91} & (25.0\%/29.3\%)\\
        Autobots + 3D Pose (arms and legs)  & 0.92/1.94 & (23.3\%/28.1\%)\\

         \bottomrule
         
    \end{tabular}
    \egroup
    \caption{Comparison between using full-body joints and selected joints based on the attention map.}
    \label{tab:selected_joints}
\end{table}

\subsubsection{Robustness in Noisy Pose}

To simulate real-world conditions where pose detection is inherently imperfect, this experiment assesses the model's robustness to inaccurate pose inputs. This evaluation is crucial for understanding how our model performs when confronted with the inaccuracies typically present in pose detection systems. For this purpose, we train the model twice: once with clean pose inputs, as used in previous sections, and once by adding Gaussian noise (mean = 0, std = 0.1) to 50\% of the scenes.
The results in \Cref{tab:add_noise} show how the models respond when Gaussian noise with zero mean and varying standard deviations is added to pose inputs during inference. Performance declines when noisy pose inputs are introduced to the model trained solely on clean data.
However, the model trained with noisy pose inputs maintains a positive gain in ADE/FDE, indicating improved robustness and reduced impact of noisy data during inference. These findings highlight the model's sensitivity to noisy data and reveal potential vulnerabilities in real-world scenarios where pose information may be less accurate.

\begin{table}[!t]
    \centering
    \bgroup
    \def\arraystretch{1.5}%
    \resizebox{.5\textwidth}{!}{
    \begin{tabular}{lccc}
    \toprule
        & \multicolumn{3}{c}{\textbf{Inference pose}} \\
        \textbf{Models} & \textbf{clean } & \textbf{noisy (std=0.2)} &  \textbf{noisy (std=0.5)} \\
        \toprule
         Autobots~\cite{girgis2022latent} & 1.20/2.70 & - & -\\
         Autobots + 3D Pose (clean) & \textbf{0.90/1.91} & 1.93/3.50  &  2.37/4.63 \\
         Autobots + 3D Pose (noisy) & 0.96/2.02 & \textbf{0.99/2.06}  &  \textbf{1.21/2.48} \\
         \bottomrule
         
    \end{tabular}
    }
    \egroup
    \caption{Comparison of the performance of models trained with trajectory only, clean pose and noisy pose on the JTA~\cite{fabbri2018learning} dataset.}
    \label{tab:add_noise}
\end{table}

\subsubsection{2D Pose vs. 3D Pose}
Until now, when we have referred to pose, we have meant 3D pose. Now, we will use the same model architecture and retrain it with 2D poses to examine the impact of using 2D poses as an alternative to 3D poses. It is worth noting that obtaining 2D pose data is generally easier than acquiring 3D pose data.
\Cref{tab:ex_jta_2d} shows that incorporating 2D pose information also improves trajectory prediction, though not as significantly as with 3D pose. This difference may be attributed to the additional depth information provided by 3D poses, which enhances the model’s understanding of spatial relationships among agents.
Despite some information loss when using 2D instead of 3D poses, our pose encoder still achieves an approximate 15\% improvement over baseline models. This demonstrates our module's effectiveness in enhancing trajectory prediction performance, even with the simpler 2D pose data.

\begin{table}[!t]  
    \centering
    \bgroup
    \def\arraystretch{1.5}%
    \begin{tabular}{lcc}
    \toprule
        \textbf{Models} &  \textbf{ADE/FDE}  & \textbf{(gain)} \\
        
        \hline
       
         Autobots~\cite{girgis2022latent} & 1.20/2.70 & - \\
         
        Autobots + 2D Pose & 1.02/2.17 & (15.0\%/19.6\%)\\
       
         Autobots + 3D Pose & \textbf{0.90/1.91} &  (25.0\%/29.3\%)\\
         \bottomrule
         
    \end{tabular}
    \egroup
    \caption{Comparison of the performance when using 2D pose instead of 3D pose on the JTA~\cite{fabbri2018learning} dataset.}
    \label{tab:ex_jta_2d}
\end{table}

\subsubsection{Robustness to Partial 2D Pose Input}

Since 2D poses are easier to capture than 3D poses, it is also important to assess the robustness of 2D pose inputs to simulate imperfect data commonly encountered in practical scenarios. The missing joints are implemented as zero-padding.
\Cref{tab:partial_pose} presents results for: (a) leg/arm occlusion with a 50\% probability, (b) structured removal of the right leg in all frames, and (c) complete-frame occlusion with a 50\% probability. The results demonstrate that the pose encoder is able to work accurately in spatial and temporal occlusion situations and the complete-frame occlusion is the most challenging case.

\begin{table}[!t]
    \centering
    \centering
    \bgroup
    \def\arraystretch{1.5}%
    
    \begin{tabular}{lc}
    \toprule
        \textbf{Inference condition}  &  \textbf{Autobots + 2D Pose}\\
        \midrule
        T + clean 2D Pose & \textbf{1.02/2.17}\\
        T + random leg and arm occlusion & 1.05/2.24\\
        T + structured right leg occlusion & 1.03/2.19\\
        T + complete-frame missing (50\%) & 1.18/2.45\\

         \bottomrule
         
    \end{tabular}
    \egroup
    \caption{Studying the effect of a partial 2D pose input on the JTA~\cite{fabbri2018learning} dataset.}
    \label{tab:partial_pose}
\end{table}

\subsubsection{Different Options to Encode and Fuse the Pose Information}
To investigate how different encoders and fusion strategies impact the performance of our pose encoder, we conduct an ablation study comparing the use of an LSTM encoder and a transformer encoder for processing pose information. After selecting the pose encoder, we further explore whether a cross-attention module that attends between trajectory embeddings and pose embeddings outperforms direct concatenation. As shown in \Cref{tab:ablation_pose_encoder}, the transformer encoder captures pose information more effectively than the LSTM encoder, and directly concatenating pose and trajectory features yields better results than using a cross-attention module. These findings highlight the effectiveness of our module’s design.

\begin{table}[!t]
    \centering
    \centering
    \bgroup
    \def\arraystretch{1.5}%
    
    \begin{tabular}{lc}
    \toprule
        \textbf{Pose encoder + fuse strategy}  &  \textbf{ADE/FDE}\\
        \midrule
        LSTM encoder + Concat. & 0.92/1.95\\
        Transformer encoder + Concat. & \textbf{0.90/1.91}\\
        Transformer encoder + Cross-att. & 0.97/2.05\\

         \bottomrule
         
    \end{tabular}
    \egroup
    \caption{Ablation study on pose encoders and fusion strategies. ``Concat'' refers to direct concatenation, while ``Cross-att.'' denotes the cross-attention module.}
    \label{tab:ablation_pose_encoder}
\end{table}

\subsection{Trajectory Prediction in 2D Pixel-space}

We further validated our method on the ACTEV benchmark~\cite{oh2011large} by comparing it against Next~\cite{liang2019peeking}, a prominent approach that also leverages multimodal inputs. We followed the experimental setup from~\cite{liang2019peeking}, using the official ACTEV dataset splits and reporting metrics for the best single prediction (Top-1) and the best of 20 samples (Top-20). As shown in \Cref{tab:results_next}, our pose-augmented Autobots model achieves better performance on both the Top-1 metric and the Top-20 metric. Specifically, during multi-plausible predictions, our method can outperform Next by up to 30\% although Next uses more input modalities. This underscores the efficacy of our approach, particularly the pose encoding component, even when using fewer input sources.

\begin{table}[t]
    \centering
    \begin{tabular}{ccc}
    \toprule
    \textbf{Model} & \textbf{Input modalities} & \textbf{MinADE$_k$/FDE$_k$}\\
    \midrule
    \multicolumn{3}{c}{Single prediction (k=1)}\\
    \midrule
    Next~\cite{liang2019peeking} & Traj.+2d P & 19.78/42.43\\
    Autobots + 2D Pose   & Traj.+2d P & \textbf{18.37}/\textbf{37.07} \\ 
    \midrule
    \multicolumn{3}{c}{Best of 20 predictions (k=20)}\\
    \midrule
    Next*~\cite{liang2019peeking} & Traj.+2d P+Activity & 16.00/32.99\\
    Autobots + 2D Pose   & Traj.+2d P & \textbf{12.43}/\textbf{22.60} \\ 
    
    \bottomrule
    \end{tabular}
    \caption{Comparison between our Pose-based Autobots and Next~\cite{liang2019peeking} on ACTEV benchmark. About input modalities, 2d P indicates 2d pose keypoints. *Results are taken from the original publication.}
    \label{tab:results_next}
\end{table}

\subsection{Pose Encoder in a Robotic Dataset}
To further investigate the model's performance in real robot scenarios, we conducted experiments on the JRDB~\cite{vendrow2023jrdb} dataset. As this dataset provides only ground truth 2D poses, we tested the pose encoder with 2D poses to show the benefit from them in real-world robotic scenarios. \Cref{tab:jrdb_2dp} shows that inputs with augmented 2D pose significantly enhance the trajectory prediction performance, with an ADE/FDE gain of up to 25\%/27\%.

\begin{table}[!t]
    \centering
    \bgroup
    \def\arraystretch{1.5}%
    \begin{tabular}{lcc}
    \toprule
        \textbf{Models} & \textbf{ADE/FDE} & \textbf{(gain)}\\
        \midrule
        Autobots~\cite{girgis2022latent} & 0.307/0.555 & -\\
        Autobots + 2D Pose & \textbf{0.230/0.405} & 25.1\%/27.0\%\\
 
         \bottomrule
         
    \end{tabular}
    \egroup
    \caption{Leverage 2D pose on the JRDB~\cite{vendrow2023jrdb} dataset. The models are trained and evaluated on samples with Trajectory and ground truth 2D pose annotations.}
    \label{tab:jrdb_2dp}
\end{table}

\subsection{Pose Encoder for Other Agents: Cyclists}

In autonomous driving applications, cyclists are also crucial participants, and the ability to predict their trajectories is necessary to provide safety. Here, we want to study the generalization ability of the pose encoder to cyclists. 
\Cref{tab:urban} compares the performance of the Autobots + 3D Pose model to the previous work~\cite{kress2022pose}, which uses 3D body poses to predict trajectories for pedestrians and cyclists. The notations `c' and `d' denote two variations of their model, using continuous or discrete approaches, respectively. The Autobots + 3D Pose model effectively leverages pose information and outperforms other models, demonstrating the effectiveness of our architecture and its capability to utilize pose data to generally improve prediction accuracy for both pedestrians and cyclists.

\begin{table}[!t]
    \centering
    \bgroup
    \def\arraystretch{1.5}%
    \begin{tabular}{lcc}
    \toprule
        \textbf{Models} &  \textbf{Pedestrians} & \textbf{Cyclists} \\
        \midrule
        c$_{traj}$~\cite{kress2022pose} & 0.57 & 0.68\\
        d$_{traj}$~\cite{kress2022pose}& 0.60 & 0.67\\
        c$_{traj,pose}$~\cite{kress2022pose}& 0.51 & 0.64\\
        d$_{traj,pose}$~\cite{kress2022pose}& 0.56 & 0.63\\

        Autobots + 3D Pose & \textbf{0.43} & \textbf{0.44}\\
         \bottomrule
         
    \end{tabular}
    \egroup
    \caption{Results on the Pedestrians and Cyclists in Road Traffic dataset~\cite{kress2022pose} in terms of ASWAEE. The lower the better.}
    \label{tab:urban}
\end{table}

\subsection{Pose Encoder to Enhance Robot Navigation}

To assess the effectiveness of our model in downstream robotic tasks, we integrate our pose-based predictor into a navigation simulation. In this simulation, a moving robot starts at an initial position and aims to reach a goal point, with the objective of doing so more quickly and with fewer collisions with neighboring agents. For evaluation, we use the completion time and collision rate used in CrowdNav~\cite{chen2019crowd} to evaluate the performance of navigation.
During the implementation, a video clip from the JTA~\cite{fabbri2018learning} test split is used to generate test trajectories, resulting in approximately 300 test samples. The simulated robot's starting and goal points are initialized as $(x_{last\_ego},y_{last\_ego}-5)$ and $(x_{last\_ego},y_{last\_ego}+5)$, where $(x_{last\_ego},y_{last\_ego})$  represents the ego agent's coordinates in the last observed frame. To integrate our predictor, we use the classic rule-based social force~\cite{helbing1995social} navigator, incorporating predicted trajectories by adding extra repulsive forces. The original rule-based navigator without trajectory prediction serves as the baseline. We then enhance the navigator by integrating Autobots and our pose-based Autobots to evaluate how trajectory predictors, particularly the pose-based version, improve navigation performance.

 \begin{table}[!t]
    \centering
    \bgroup
    \def\arraystretch{1.5}%
    \resizebox{.5\textwidth}{!}{
    \begin{tabular}{lcc}
    \toprule
        \textbf{Navigation} & \textbf{Completion time $\downarrow$ (degradation)} & \textbf{Collision rate $\downarrow$ (degradation)}\\
        
        \midrule
        w/o trajectory prediction  & 13.86 & 6.60\%\\
        w/ Autobots~\cite{girgis2022latent} & 13.27 (4.3\%) &  5.56\% (15.8\%) \\
        w/ Autobots + 3D Pose & \textbf{12.63} (8.9\%) & \textbf{4.17\%} (36.8\%)\\

        \bottomrule
         
    \end{tabular}
    }
    \egroup
    \caption{Quantitative results of the robot navigation task. The completion time is reported in seconds and collision rate is reported in percentage.}
    \label{tab:navigation}
\end{table}

\begin{figure}[!t]
\centering
\begin{subfigure}{.5\textwidth}
    \hspace{.15\linewidth}
    \begin{frame}{}
    \animategraphics[loop,controls,poster=20,width=0.65\linewidth]{10}{figures/navigation/baseline/frame_}{00}{20} 
    \end{frame}
    \caption{Robot navigation without Social-pose. We observe that a collision could happen as the robot cannot effectively predict others.}\label{fig:navigation_a}
  \end{subfigure}%\\
\hfill
  \begin{subfigure}{.5\textwidth}
\hspace{.15\linewidth}
  \begin{frame}{}
  \animategraphics[loop,controls,poster=18,width=0.65\linewidth]{10}{figures/navigation/predictor/frame_}{00}{32} 
    \end{frame}
    \caption{Robot navigation with Social-pose. We observe that the robot could avoid collision by using our pose-based predictor.} \label{fig:navigation_b}
  \end{subfigure}%

\caption{Qualitative results of the robot navigation task without (on top) and with (on bottom) social-pose. It is best viewed using Adobe Acrobat Reader.} \label{fig:navigation}
\end{figure}

\Cref{tab:navigation} presents the quantitative results of applying our method to robotic navigation tasks. The results show a reduction in completion time and collision rate by approximately 9\% and 37\%, respectively. \Cref{fig:navigation} qualitatively illustrates a scenario where the robot successfully bypasses pedestrians earlier to avoid collision by incorporating the predicted future trajectories of nearby pedestrians. Our experiments showed that incorporating the pose-based trajectory predictor enabled the robot to reach its goal more quickly and with a lower collision rate.

\section{Conclusion}
We have proposed Social-pose, a lightweight decoupled pose encoder that captures spatiotemporal interactions between pedestrians by attending to body poses. Through extensive experiments, we have demonstrated that incorporating pose information can significantly enhance the performance of various models, including LSTM, GAN, MLP, and Transformer-based architectures. Moreover, we explored the effects of 2D vs. 3D poses and the effect of noisy pose data on the task, as well as the benefits of our pose-based predictors in robot navigation scenarios.

While our proposed attention-based encoder is generic, some applications might suggest using a sparse number of keypoints. In future work, one can explore ways to extract more compact and relevant information from poses, acknowledging that considering the entire set of keypoints might not always be necessary.
Yet, it might be quite application-specific.

\bibliographystyle{ieee_fullname}
\bibliography{refs}

\newpage
\begin{IEEEbiography}
[{\includegraphics[width=1in,clip,keepaspectratio]{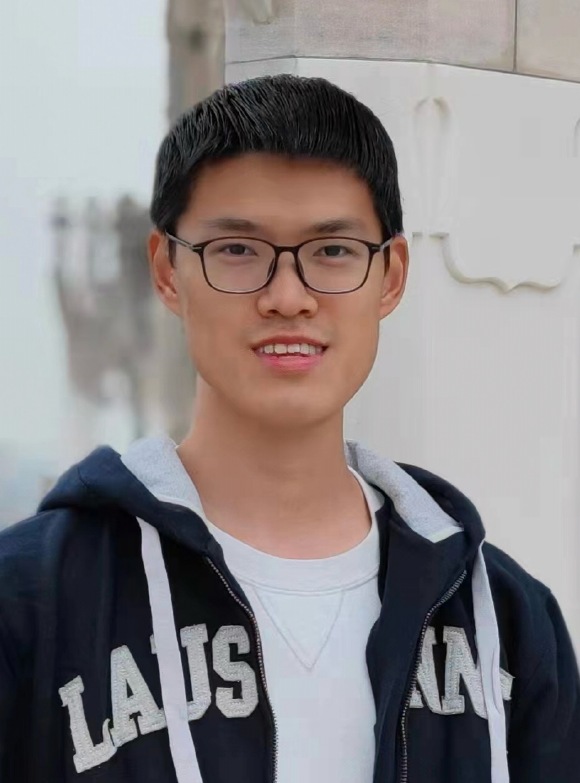}}]%
{Yang Gao}
received the B.E. degree from Shanghai Jiao Tong University, China, in 2021, and the M.S. degree from KTH Royal Institute of Technology, Sweden, in 2022. Currently, he is pursuing the Ph.D. in the Visual Intelligence for Transportation Laboratory (VITA) at EPFL. His research interests include human trajectories and pose keypoints forecasting in a variety of robotic and traffic scenarios.

\end{IEEEbiography}

\begin{IEEEbiography}[{\includegraphics[width=1in,clip,keepaspectratio]{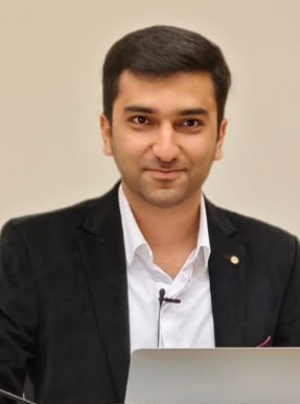}}]%
{Saeed Saadatnejad} is a research scientist at EPFL, where he earned his Ph.D. in computer science. He was awarded the EPFLInnovators fellowship under the Marie-Curie grant for his doctoral degree.
Previously, he received his BSc and MSc from Sharif University of Technology in 2015 and 2018, respectively. 
His research interests include deep generative models and motion / behavior prediction.

\end{IEEEbiography}

\begin{IEEEbiography}[{\includegraphics[width=1in,clip,keepaspectratio]{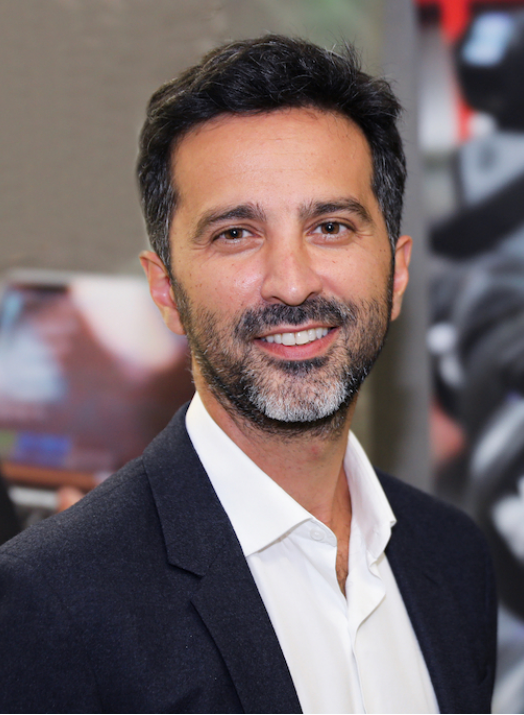}}]%
{Alexandre Alahi}  (Member, IEEE) is currently an
Associate Professor with EPFL, where he is leading
the Visual Intelligence for Transportation Laboratory
(VITA). Before joining EPFL in 2017, he spent
multiple years as a Post-Doctoral Researcher and
a Research Scientist at Stanford University. His
research interests include computer vision, machine
learning, and robotics applied to transportation and
mobility.

\end{IEEEbiography}

\end{document}